\pdfoutput=1

\documentclass[11pt]{article}

\usepackage{acl}

\usepackage{times}
\usepackage{latexsym}

\usepackage[T1]{fontenc}

\usepackage[utf8]{inputenc}

\usepackage{microtype}

\usepackage{amsmath,amsfonts,bm, bbm}
\usepackage{dblfloatfix}
\usepackage{booktabs}

\def\eqref#1{equation~\ref{#1}}

\def\1{\bm{1}}

\DeclareMathAlphabet{\mathsfit}{\encodingdefault}{\sfdefault}{m}{sl}
\SetMathAlphabet{\mathsfit}{bold}{\encodingdefault}{\sfdefault}{bx}{n}

\DeclareMathOperator*{\argmax}{arg\,max}

\newcommand{\newtext}[1]{#1}

\usepackage{booktabs, multirow, multicol}
\usepackage{hyperref}
\usepackage{url}
\usepackage{wrapfig}
\usepackage{graphicx}
\usepackage{caption}
\usepackage{subcaption}

\title{Downstream Trade-offs of a Family of Text Watermarks}

\author{Anirudh Ajith\thanks{~~Work done during an internship at IISc, Bangalore.} \\
  Princeton University \\
  \href{mailto:anirudh.ajith@princeton.edu}{anirudh.ajith@princeton.edu}
  \\\And
  Sameer Singh \\
  University of California, Irvine \\
  \href{mailto:sameer@uci.edu}{sameer@uci.edu}
 \\\And
  Danish Pruthi \\
  Indian Institute of Science\\ 
 \href{mailto:danishp@iisc.ac.in}{danishp@iisc.ac.in}
  }

\begin{document}
\maketitle

\begin{abstract}
Watermarking involves implanting an 
imperceptible signal into 
generated text that can later be
detected via statistical tests. 
A prominent family of watermarking strategies 
for LLMs 
embeds this signal by 
upsampling a  (pseudorandomly-chosen) 
subset of tokens at every generation step. 
However, such signals alter the model's
output distribution 
and can have unintended effects 
on its downstream  performance. 
In this work, 
we evaluate the performance of 
LLMs 
watermarked  
using 
three different strategies 
over a diverse suite of tasks
including those cast as $k$-class classification (CLS), multiple choice question answering (MCQ), 
short-form generation (e.g., open-ended question answering) and long-form generation (e.g., translation) tasks.
We find that watermarks 
(under realistic hyperparameters) 
can cause significant drops in LLMs' 
effective utility across all tasks. 
We observe drops of $10$ to $20$\% in CLS tasks in the average case, which shoot up to $100$\% in the worst case. 
We notice degradations 
of about $7$\% in MCQ tasks, $10$--$15$\% in short-form generation, and $5$--$15$\% in long-form generation tasks.
Our findings highlight the trade-offs that users should be cognizant of when using watermarked models.\footnote{We make our code available at \url{https://github.com/FLAIR-IISc/watermark_tradeoffs}.}
\end{abstract}

\section{Introduction}

Large Language Models (LLMs), and derived chatbots of the likes of ChatGPT, can generate human-like responses to a variety of requests like writing emails, translating or summarizing content~\citep{brown2020language, chowdhery2022palm}. 
As these systems gain popularity, there are looming concerns about their misuse 
for spreading targeted misinformation, 
influencing public opinion~\citep{panditharatne2023elections} or conducting social engineering attacks~\citep{grbic2023social}. 

Such concerns have spurred research towards 
distinguishing human-written and LLM-generated content. 
Naive approaches such as 
training post-hoc classifiers 
for this purpose 
have been shown to be ineffective,
as they 
typically have large false-positive rates 
that can lead to 
false accusations of plagiarism~\citep{O’Neill_2023, OpenAI2023detection}. 
These classifiers 
can further degrade in accuracy when LLM developers like OpenAI continually finetune and update their public models. Additionally, 
the output distributions of future LLMs 
may grow even more similar to that 
of human-written text 
causing the efficacy of such approaches to wane.

A promising alternative is to intentionally embed a \textit{watermark} signal~\citep{atallah2001natural,chiang2004natural, topkara2006natural, jalil2009review} into LLM-generated text
that is imperceptible to unsuspecting readers 
but can be algorithmically detected using statistical tests. A popular watermarking scheme, often referred to as KGW, works by boosting the probabilities of a psuedorandomly chosen subset of the model's vocabulary at every generation step~\cite{pmlr-v202-kirchenbauer23a}. This scheme has been extensively studied and extended~\citep{kirchenbauer2023reliability, liu2023semanticrobust4, lu-etal-2024-entropy}. The original approach and its derivatives, collectively called the KGW family in the literature, comprise the most popular watermarking strategies for LLMs today.

Previous works studying the trade-offs of watermarking LLMs mostly restrict their analysis to intrinsic evaluations of watermarked models generation quality such as perplexity or GPT4 judgements~\citep{singh2023new}, eschewing evaluation on downstream task benchmarks. But since it is likely that all strong LLMs made available to the public (eg. through internet APIs) will be watermarked in the near future 
(as promised by several leading LLM developers~\citep{whitehouse2023press}), it is important to understand how watermarks impact LLMs' performance on downstream tasks.

In our work, we evaluate the downstream impact of 3 popular watermarks from the KGW family including the original KGW approach~\citep{pmlr-v202-kirchenbauer23a}, EWD~\citep{lu-etal-2024-entropy} and SIR~\citep{liu2023semanticrobust4} over a diverse selection of tasks. Since KGW-based watermarks perturb output probability distributions at the token level, we categorize the tasks as follows for our analysis:
\begin{enumerate}
    \item CLS: Tasks framed as $k$-class classification problems with static labels. 
    \item MCQ: Tasks framed as multiple choice question-answering problems with  choices that differ across test examples.
    \item SGEN: Tasks requiring generation of a short output sequence via sampling. %
    \item LGEN: Tasks involving the generation of a long output sequence by repeatedly sampling from the LLM's probability distributions.
\end{enumerate}

We categorize the examined tasks into these four buckets as we expect similar effects of watermarking for tasks in a given category. 
For instance, 
for CLS tasks, there is a possibility 
of systematic bias against some labels for every test example. 
In SGEN and LGEN tasks, 
the modifications to the output distributions  
due to watermarking can  
directly impact the correctness of generated content.\footnote{We treat short-form and long-form generation tasks differently due to differences in how they are evaluated.}

We evaluate the performance of watermarked LLaMA~\citep{touvron2023llama}, Mistral~\citep{jiang2023mistral} and OPT~\citep{zhang2022opt} models and observe that, under realistic watermark settings, watermarking can cause significant drops in LLMs' effective utility across all tasks. 
We notice drops of $10$--$20$\% in CLS tasks in the average case which can rise up to $100$\% in the worst case. 
We see drops of about $7$\% in MCQ tasks, $10$--$15$\% in short-form generation, and $5$--$15$\% in long-form generation.%

We believe that our findings will allow model developers and users to make informed choices about watermarked models and spur interest into developing novel watermarking schemes and decoding strategies that may exhibit better performance trade-offs. We make our code available at \url{https://github.com/FLAIR-IISc/watermark_tradeoffs} to facilitate research in this area, and holistically evaluate future watermarking approaches.

\section{Background}
\label{sec:background}

The KGW watermark~\citep{pmlr-v202-kirchenbauer23a} is a deterministic algorithm parameterized by 3 hyperparameters $\gamma, \delta$ and $k$, and a keyed psuedorandom function $F_\cdot : \mathbb{N} \rightarrow \{g, r\}^m$. 

\paragraph{Generation.} 
The algorithm works by modifying the logits obtained from the language model at each generation step. Formally, given a model $\mathcal{M}$ with vocabulary $V$, and a prefix comprising tokens $\mathbf{w}_1, \mathbf{w}_2, \ldots, \mathbf{w}_n$ the scheme involves first computing the logits $\mathcal{M}(\mathbf{w}_1 \ldots, \mathbf{w}_n) = (l_1, \ldots, l_{|V|})$ of the language model that would ordinarily be used to predict the %
subsequent token. The terminal prefix token $\mathbf{w}_n$ is then fed to $F$ under the key $k$ to obtain a partition of $V$ into a green list $G$ and a red list  $R$ such that $|G| = \lfloor \gamma |V| \rfloor$. That is,
$$F_k(\mathbf{w}_n) \in \{g, r\}^{|\mathcal{V}|}$$
such that $\sum_{x \in F_k(\mathbf{w_n})} \mathbbm{1} [x = g] = \left \lfloor \gamma |V| \right \rfloor.$ 
Finally, watermarked logits $(\lambda_1, \ldots, \lambda_{|V|})$ are computed as $\lambda_i = l_i + \delta \cdot \mathbbm{1} [i \in G]$. These watermarked logits can then be used for sampling tokens (for generation) or even computing the likelihood or perplexity of a given sequence.

\paragraph{Detection.} 
The detection scheme proposed by \citet{pmlr-v202-kirchenbauer23a} works by assessing the probability of a null hypothesis that the given text was written without knowledge of the watermarking scheme (specifically hash key $k$). Precisely, given a token sequence $x$ of length $T$ that was written without knowledge of the scheme, the number of green list tokens in $x$, denoted by $|x|_G$, can be assumed to be normally distributed with a mean of $\gamma T$ and a standard deviation of $\sqrt{T\gamma(1-\gamma)}$. The detection algorithm computes a z-score, 
\begin{equation}
\label{eqn:zscore}
z = (|x|_G - \gamma T) / \sqrt{T\gamma(1-\gamma)}, 
\end{equation} and rejects the null hypothesis if this z-score exceeds a chosen threshold.

\paragraph{Variations.} 

Entropy-based Watermarking Detection~(EWD) is a variation of the KGW approach that uses the same generation algorithm but differs in its detection algorithm~\citep{lu-etal-2024-entropy}. EWD seeks to improve the trade-off between watermark detectability and language modeling ability by proposing a novel entropy-based detection strategy that involves reweighting individual tokens using their entropies during detection. It computes an adjusted z score as
$$z' = (|x|_G - \gamma \sum_{i=m}^{|T|-1}W_i) / \sqrt{T\gamma(1-\gamma) \sum_{i=m}^{|T|-1}W_i^2}$$
where $W_i$ is computed based on the entropy of the $i^\text{th}$ token of the sequence $x$.

Another KGW-based derivative called Semantic-Invariant and Robust (SIR) watermarking aspires to improve the robustness against paraphrasing attacks \citep{liu2023semanticrobust4}. The SIR watermarking scheme computes $G$ as a pseudorandom function of a semantic embedding of its prefix, departing from KGW's strategy of computing $G$ using the prefix's lexical properties. For SIR, 
\begin{align*}
F'_k(\text{embed}(\mathbf{w}_1 \ldots, \mathbf{w}_n)) = \{g, r\}^{|V|}    
\end{align*}
Specifically, SIR utilizes a sentence encoder to generate a semantic embedding for the prefix and then uses a learned `watermark model' to transform this embedding into a partition over the model's vocabulary. A notable feature of the SIR watermark is that the $\gamma$ hyperparameter cannot be set explicitly, but instead is implicitly determined by the watermark model at every generation step. This model's training objective incentivizes the effective $\gamma$ to always take on values close to $0.5$. 

\section{Evaluation Setup}
\label{sec:evaluation_approach}

\label{sec:evaluation_approach_setup}
\paragraph{Datasets.} 
\newtext{To assess watermarks' effects} on tasks that are framed as classification tasks (CLS), we work with SST-2, BoolQ and CB from the GLUE~\citep{wang2019glue} and SuperGLUE~\citep{wang2019superglue} benchmarks. These correspond to sentiment analysis, yes/no question answering and textual entailment tasks respectively. We select the commonsense NLI dataset called HellaSwag \citep{zellers2019hellaswag} and the question-answering dataset PIQA~\citep{Bisk2020PIQA} as MCQ tasks, the reading comprehension datasets DROP~\citep{Dua2019DROP} and SQuADv2~\citep{2016arXiv160605250R} as SGEN tasks, and the WMT14-En-Fr~\citep{bojar-EtAl:2014:WMT14} and WMT20-En-De~\citep{wmt20barrault-etal-2020-findings} translation tasks as LGEN tasks. We evaluate models' performance on these tasks using the metrics \newtext{typically} associated with them. For instance, CLS and MCQ tasks are evaluated using accuracy while SGEN tasks are evaluated using F1 scores and LGEN translation tasks are evaluated using BLEU scores~\citep{papineni2002bleu}.
\noindent
\begin{figure*}[ht]
     \centering
     \includegraphics[width=\textwidth]{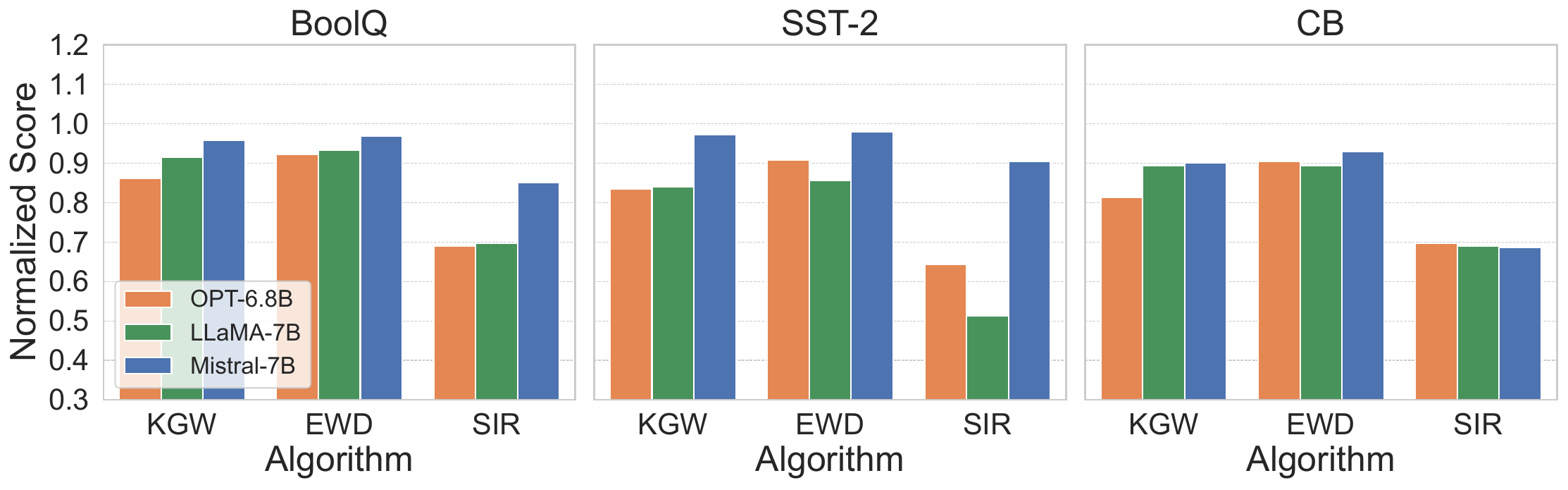}
     \caption{Expected normalized scores for classification datasets. KGW and EWD show $10$ - $20$\% drops in normalized scores while while SIR can cause drops of $30$ - $60$\%.}
     \label{fig:classification} 
\end{figure*}

\paragraph{Models.} We analyze the performance trade-offs for the above  tasks for watermarked and unwatermarked \newtext{versions of} LLaMA 7B~\citep{touvron2023llama}, Mistral 7B~\citep{jiang2023mistral} and OPT 6.8B~\citep{zhang2022opt} models. 

\paragraph{Methodology.} In KGW-based watermarks, the $\gamma$ and $\delta$ hyperparameters control the strength of the watermark signal and accordingly, the shift in watermarked models' output distribution. However due to differences among these algorithms, a specific $(\gamma, \delta)$ setting does not imply the same signal strength (as measured by its empirical detectability) or impact on an LLM's language modeling ability.

To ensure a fair comparison of these schemes' downstream implications, 
we find the settings of hyperparameters for each watermarked model such that the resulting signal is of the same strength. 
In the watermarking literature, signal strengths are typically  evaluated by  
computing the True Positive Rates (TPR) of their detection algorithm at a fixed False Positive Rate (FPR). Generally, FPR is set to a low value such as $0.01$, to avoid the risk of false accusing someone of plagiarism.
In our evaluation, we consider
signal strengths of $0.5$, $0.75$ and $0.95$ TPR@FPR=$0.01$ at $50$ generated tokens to be \textit{light}, \textit{moderate} and \textit{heavy} intensity settings respectively. 

For each watermark and model, we use 200 prefixes sampled from the C4 corpus~\citep{2020t5} as prompts to isolate the $\delta$ values corresponding to light, moderate and heavy watermarks for each $\gamma \in \{0.1, 0.25, 0.5, 0.75\}$. Next, we obtain the perplexity values for each $(\gamma, \delta)$ setting corresponding to a \newtext{particular} signal strength over \newtext{a disjoint} sample of $200$ C4 snippets. We \newtext{then} choose the \newtext{tuple} which least impacts the model's perplexity \newtext{scores} as the canonical hyperparameter setting for that signal strength. 
Through this process, we select the pareto-optimal set of hyperparameters with respect to language modeling performance under a target \newtext{watermark strength}. 
Some contemporary work~\citep{tu-etal-2024-waterbench} that performs downstream evaluations fails to 
conduct this type of pareto-optimal hyperparameter search, 
and instead arbitrarily chooses a $(\gamma, \delta)$ setting that achieves the target signal strength. 
We believe that this limits the practical applicability of their findings. 

\paragraph{Measuring effective utility drop.} We work under the assumption that the metrics (e.g., accuracy, F1, BLEU, etc.) that are typically used to evaluate these tasks are representative of human perception of performance on them. For CLS and MCQ tasks, even a random classifier achieves a non-zero accuracy in expectation while a random generator achieves negligible F1 or BLEU scores on the generation tasks we study. 
To measure effective utility, we compute how much the performance metric  exceeds that of a random classifier/generator. Specifically, we report normalized scores for each task in section~\ref{sec:results}. A normalized score of 0 indicates performance equivalent to that of a random classifier/generator and a normalized score of 1 indicates performance equivalent to the corresponding unwatermarked model. Specifically if $\mathcal{M}$ is an unwatermarked model, $\mathcal{M}_{w,i}$ is the corresponding model watermarked using watermark $w$ at signal strength $i$, if $t$ is a downstream task and if $\mathcal{M}(t)$ denotes model's raw score on the task,
then 
\begin{align*}
\text{normalized score}(\mathcal{M}, w, i, t) = \frac{\mathcal{M}_{w,i}(t) - \mathcal{R}(t)}{\mathcal{M}(t) - \mathcal{R}(t)}    
\end{align*}
where $\mathcal{R}(t)$ indicates random performance on $t$. Note that this quantity can be negative if $\mathcal{M}_{w,i}(t) < \mathcal{R}(t) < \mathcal{M}(t)$. \newtext{It can also exceed $1$ if $\mathcal{M}_{w,i}(t) > \mathcal{M}(t)$.}

\section{Results}
\label{sec:results}
We present our main results over all models, tasks and watermarks under the \textit{moderate} signal strength and provide our full set of results in Table~\ref{tab:all_scores} in Appendix~\ref{app:all_scores}.

\subsection{CLS tasks}
When a fixed prompt template is used to prompt LLMs to solve classification tasks (where all test examples share a common label set), a systematic bias over the tokens comprising the label set can arise at the label generation position, and this bias can persist over all test examples. 
This could be true in the case of KGW and EWD if the terminal tokens in the prompt template (such as \texttt{Answer:}) remain fixed and in case of SIR if the semantic embedding of the prompt turns out to be similar for all test examples (e.g., on using fixed instructions and few-shot demonstrations). How specifically the label set tokens get segregated into green list $G$ and red list $R$ however, depends on the choice of the hash key $k$ used with $F$.

Using the logits of the unwatermarked models and knowledge of the tokens comprising the labels for each CLS task, we compute the expected classification task score under a uniform choice of $k$ for each watermarked model. In Figure~\ref{fig:classification}, we show that expected normalized scores can drop by at least $10$--$20$\% in CLS tasks as is the case for all models under the KGW and EWD watermarks. However, these drops can be as high as $30$--$50$\% for some models under the SIR watermark. We provide some intuition for this \newtext{discrepancy} in Section~\ref{sec:analysis}.
\noindent
\begin{figure*}[ht]
     \centering
     \includegraphics[width=\textwidth]{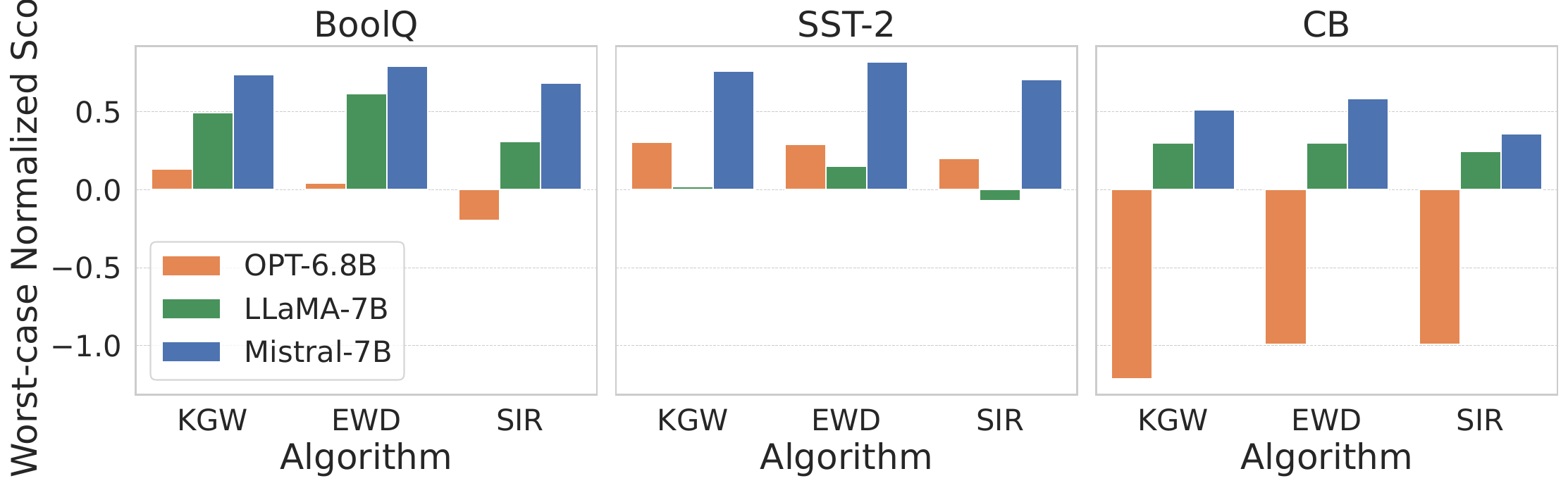}
     \caption{Normalized scores for classification datasets under the worst-case partition. For the $3$ watermarking variations, we find that the effective utility of a model can be nearly or completely lost even by moderate watermarks.}
     \label{fig:classification_worst_case}
\end{figure*}

\noindent
\begin{figure*}[ht]
    \centering
    \begin{subfigure}[b]{0.42\textwidth}
        \includegraphics[width=\linewidth]{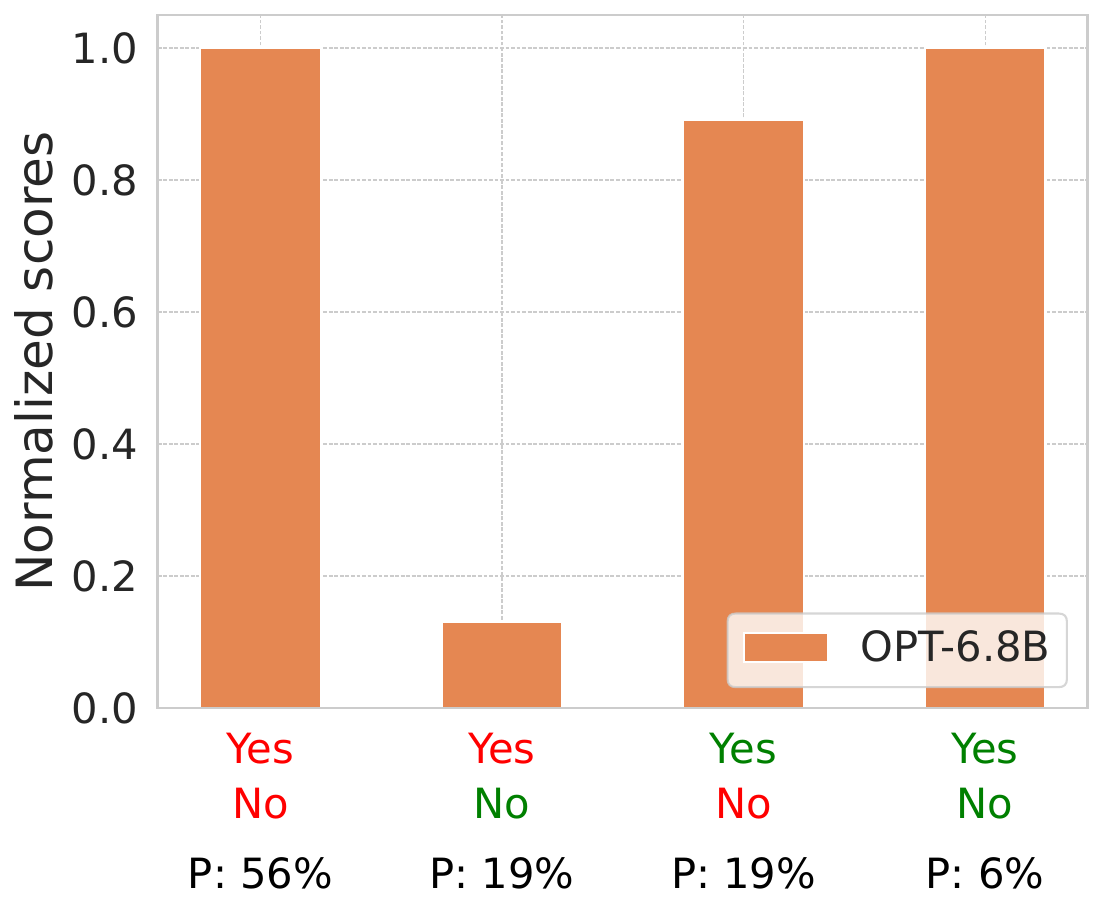}
        \caption{BoolQ}
        \label{fig:boolq_cases}
    \end{subfigure}
    \hfill
    \begin{subfigure}[b]{0.57\textwidth}
        \includegraphics[width=\linewidth]{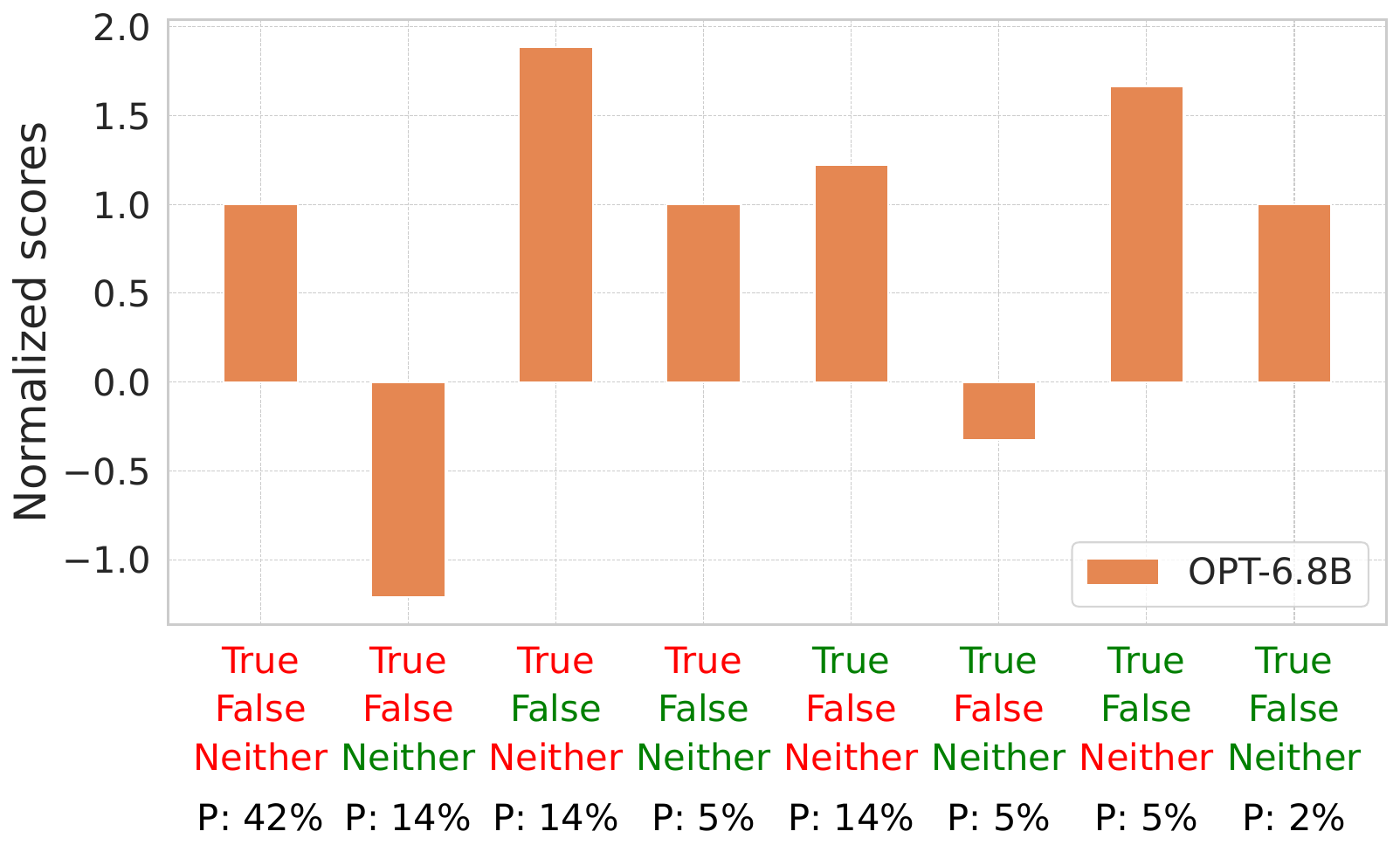}
        \caption{CB}
        \label{fig:cb_cases}
    \end{subfigure}
    \caption{OPT 6.8-B's normalized scores on BoolQ and CB under various partitions of the corresponding class labels. Labels are partitioned differently with significant probability. Effective utility of the model can be completely destroyed even under moderate watermarks.}
    \label{fig:boolq_and_cb_all_cases}
\end{figure*}

If the label set consists of $L$ tokens for a given task, then there could be $2^{|L|}$ possible partitions of this set into $G$ and $R$. By enumerating these partitions and evaluating the watermarked model under each of these partitions independently, we isolate the partition that yields the worst test accuracy and plot the corresponding worst-case normalized scores in Figure~\ref{fig:classification_worst_case}. We see that even watermark signals of moderate strength can destroy effective utility with normalized scores dropping to near zero i.e. akin to random classifier performance in OPT and LLaMA. In tasks such as CB dataset where there is a class imbalance (the minority class, \texttt{Neither}, constitutes only $6$\% of test examples), placing the minority class token into $G$ and the rest into $R$ causes accuracy to fall far below random.

It is crucial to note that these types of worst-case partitions are not rare (under a uniformly random choice of $k$). In Figure~\ref{fig:boolq_and_cb_all_cases}, we show the impact of watermarking OPT-6.8B when evaluating over the BoolQ and CB datasets. We show all possible partitions of $L$ into $G$ and $R$, and the probabilities of such partitions occurring under a randomly chosen hash key. Since these labels are only single token long, $|L| \ll |V|$, we can approximate the probability of a given partition using the binomial distribution as $$\binom{{|L|}}{n_G} \cdot \gamma^{n_G} \cdot (1- \gamma)^{|{L}| - n_G}$$ where $n_G = |{L} \, \cap \, G|$. We find that moderate watermarks can reduce BoolQ accuracy to that of a random classifier with an $19\%$ probability and the normalized score for the CB task to near-zero with a similar probability. Also notice that boosting the logits of the minority class (i.e., \texttt{Neither}) without modifying the logits of the remaining two classes causes accuracy to drop far below the random guessing baseline. Imporantly, these observations about worst case performance are consistent over all 3 the KGW-based watermarks we study.

\noindent
\begin{figure*}[ht]
     \centering
     \includegraphics[width=\textwidth]{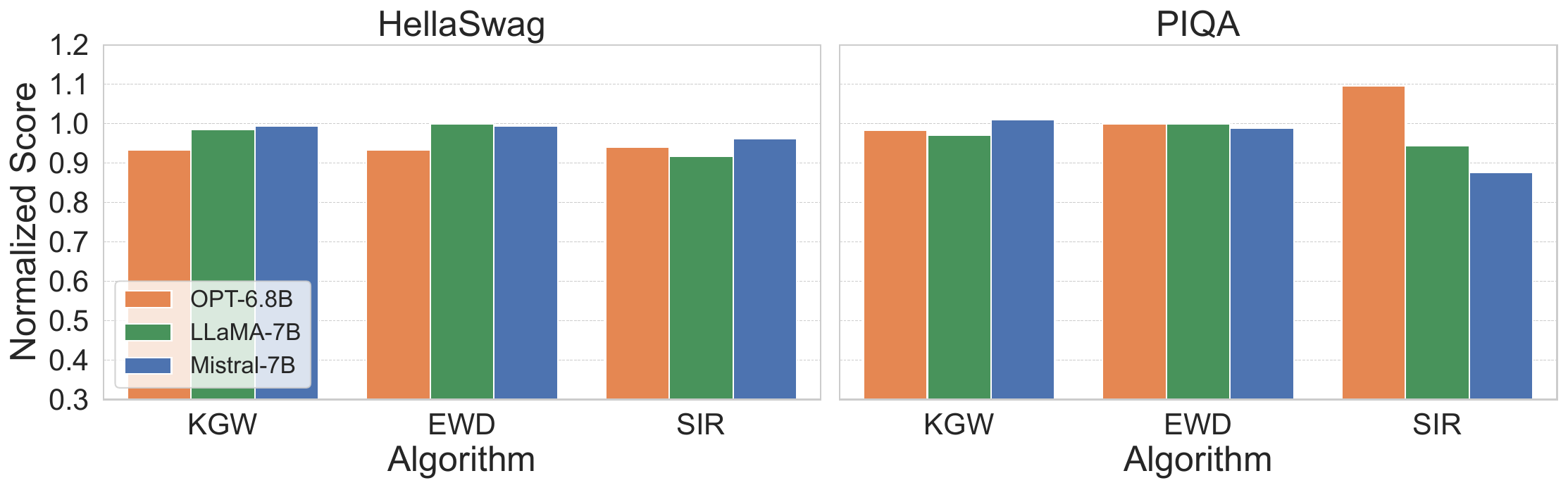}
     \caption{Expected normalized scores for multiple-choice question-answering datasets. These tasks are only mildly affected by watermarks with drops underate moderate watermarks usually restricted to $\le 10\%$.}
     \label{fig:mcq}
\end{figure*}

\noindent
\begin{figure}
    \centering
    \includegraphics[width=\linewidth]{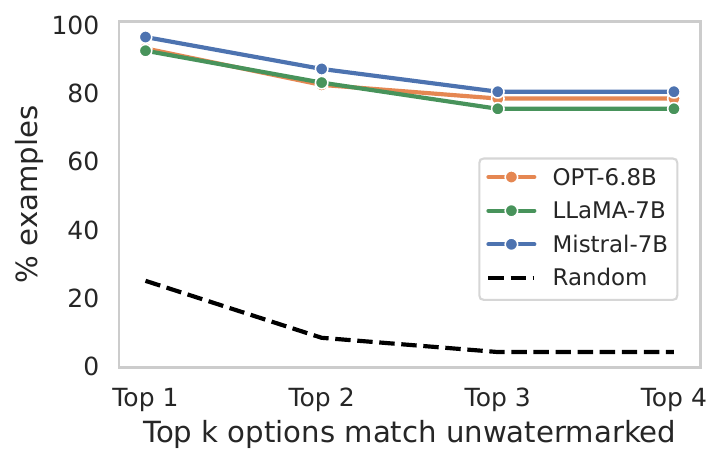}
    \caption{Proportion of HellaSwag examples where the watermark does not change the ranking of the model's top $k$ preferred options. These proportions are consistently higher than expected from a random permutation.}
    \label{fig:hellaswag_match_indices}
\end{figure}
\noindent
\begin{figure}
    \centering
    \includegraphics[width=\linewidth]{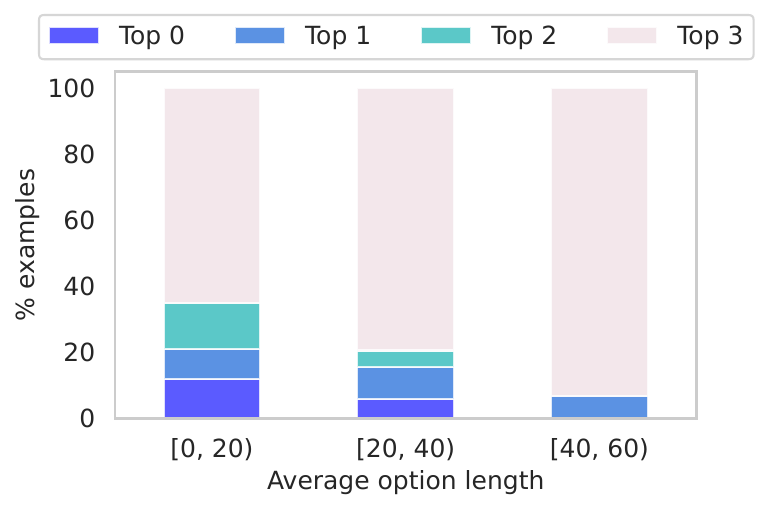}
    \caption{Proportion of HellaSwag examples where the model's top $k$ most preferred choices remains unchanged on watermarking, stratified by average option length. Rankings of examples with longer options are more robust to watermarking.}
    \label{fig:hellaswag_match_vs_length}
\end{figure}

\subsection{MCQ tasks}

We observe that watermarks leave model performance on MCQ tasks relatively unaffected. We usually see only $5$--$10$\% drops in normalized scores (Figure~\ref{fig:mcq}) which is markedly lower than those observed in CLS tasks (Figure~\ref{fig:classification}). In fact, Figure~\ref{fig:hellaswag_match_indices} shows that the model's preference ranking over all provided choices often remains unchanged by the watermark. We plot the proportion of examples from the HellaSwag task where the model's preference order over its top $k$ 
most preferred choices remains unchanged upon watermarking. Although these proportions drop (as expected) on increasing $k$, they  remain significantly higher than would be expected from a random permutation.

This discrepancy in the score drops we observe between CLS and MCQ tasks
could be due to the fact that systematic biases present in CLS tasks with static label sets do not appear in MCQ tasks (where labels differ across examples). Such an effect, however, would diminish on averaging performance over multiple hash keys (as we do). This observation is instead due to the significantly longer choice lengths in HellaSwag and PIQA (averaging $34$ and $17$ words respectively). In a longer token sequence, the fraction of tokens that occur in $G$ is more likely closely approximated by $\gamma$. Hence, each choice sees an almost uniform increase in perplexity on KGW-based watermark application, leaving the ordinal relationship among the the choices' perplexities unchanged.  In Figure~\ref{fig:hellaswag_match_vs_length}, we bucket HellaSwag examples by their average choice length (in words) and show the proportion of examples in each bucket whose top $k$ most preferred choices (taken together) remains unchanged on watermarking. Notice that the rankings of examples with longer options are more robust. %

\noindent
\begin{figure*}[ht]
     \centering
     \includegraphics[width=\textwidth]{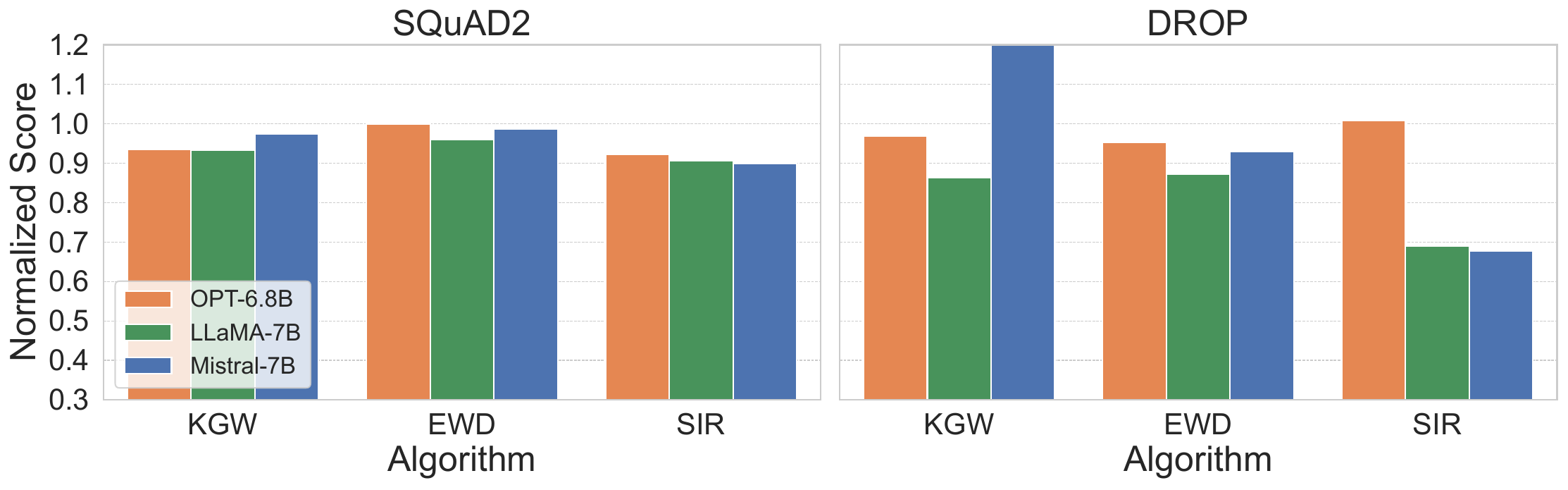}
     \caption{Expected normalized scores for short-form generation (SGEN) tasks.}
     \label{fig:shortgen}
\end{figure*}

\noindent
\begin{figure*}[ht]
     \centering
     \includegraphics[width=\textwidth]{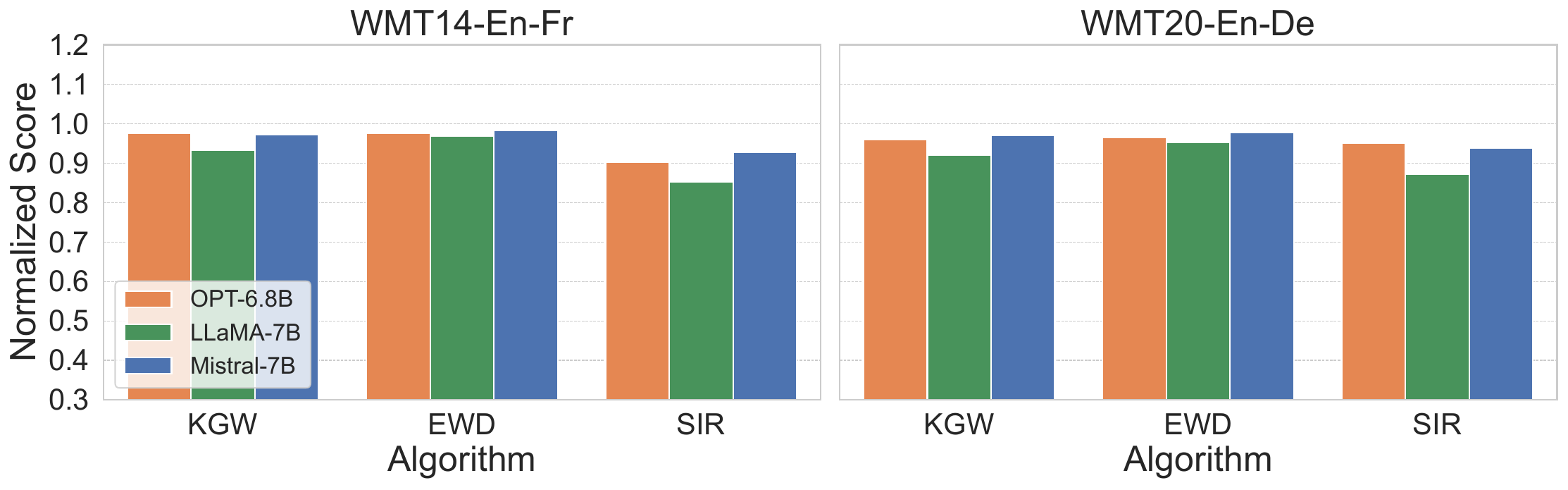}
     \caption{Expected normalized scores for long-form generation datasets. The normalized BLEU scores for both translation tasks remain almost unaffected by moderate KGW-based watermarks, with drops of about $5 - 10\%$.}
     \label{fig:longgen}
\end{figure*}

\subsection{*GEN tasks}
\label{sec:results_generation}

\begin{figure}[h]
    \centering
    \includegraphics[width=0.8\linewidth]{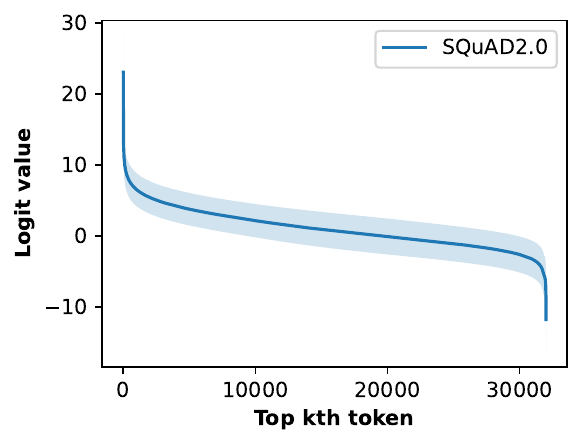}
    \caption{LLaMA 7B's logit-magnitudes for the top-$k$th token sorted in descending order, averaged over outputs at every generation step for SQuAD2 examples. We observe that at each generation step, a few tokens have significantly higher values than most others.}
    \label{fig:squad2_logits}
\end{figure}

\paragraph{SGEN} Figure~\ref{fig:shortgen} shows that models' normalized F1 scores drop by up to $10\%$ across all watermarks in SQuAD2 and by upto $15$\% on DROP. This impact is noticeable, but milder than might be expected for short generation lengths (considering our previous findings). However, it is unsurprising since in (unwatermarked) LLMs, the logits of the model's predictions at a generation step typically far exceed the logits over most other vocabulary tokens (example in Figure~\ref{fig:squad2_logits}). Since we use greedy sampling to elicit a model's predictions on SGEN tasks, a watermark would only change these predictions when $\delta$ is large enough to boost the logits of arbitrary tokens in $G$ to values larger than those of the original prediction. This appears to only rarely occur in most cases we evaluate. SIR again appears disproportionately impacted, especially in the DROP task. %

\noindent

\paragraph{LGEN} The models we study show normalized BLEU score drops of $5$--$10$\% under KGW and EWD watermarks, and up to $15$\% under SIR watermarks. We also observe qualitatively that moderate KGW-based watermarks can lead to occasional factual errors in model generations (examples in Tables~\ref{tab:enfr_translation_examples} and \ref{tab:ende_translation_examples} in Appendix~\ref{app:results}). %

\section{Analysis}
\label{sec:analysis}
\paragraph{Why does SIR perform worse?} In Section~\ref{sec:results}, we evaluate the performance drops on various different tasks due to applying moderate strength KGW, EWD and SIR watermarks over 3 models, each of about 7 billion parameters. As explained in Section~\ref{sec:evaluation_approach}, we define the `moderate' intensity as the watermark strength required to obtain an empirical TPR @ FPR $= 0.01$ of $0.75$ using $50$-token long freeform generations. Despite this uniformity in calibration, we find that the SIR watermark consistently leads to worse downstream tradeoffs than KGW and EWD over each CLS (Figure~\ref{fig:classification}), MCQ (Figure~\ref{fig:mcq}), SGEN (Figure~\ref{fig:shortgen}) and LGEN (Figure~\ref{fig:longgen}) task we study. 

One aspect that makes SIR different from KGW and EWD is that its $\gamma$ hyperparameter cannot be explicitly set by the user, but instead varies dynamically around $0.5$. As Table~\ref{tab:gamma_delta} in the appendix shows however, the $\gamma$ settings for KGW and EWD we obtain through the calibration procedure described in Section~\ref{sec:evaluation_approach} are all either $0.1$ or $0.25$. \newtext{Noticing that the likelihood of an arbitrary (independently chosen) pair of tokens being segregated differently into $G$ and $R$ is given by $1 - \gamma^2 - (1-\gamma)^2$ and that this expression is maximized at $\gamma=0.5$, we hypothesize that SIR's systematic underperformance may be due to the fact that $\gamma=0.5$ ``maximally" perturbs the model's output distribution. To verify this claim, we evaluate the performance drops due to KGW and EWD when $(\gamma, \delta)$ are set to the values that realize the `moderate' intensity when $\gamma=0.5$.}

\begin{table}[ht]
\centering
\resizebox{0.49\textwidth}{!}{  
\begin{tabular}{lccc|cc}
\toprule
 & \multicolumn{3}{c}{\textbf{moderate $\gamma$}} & \multicolumn{2}{c}{\textbf{$\gamma=0.5$}} \\
\cmidrule(lr){2-4} \cmidrule(lr){5-6}
 \textbf{Dataset} & \textbf{KGW} & \textbf{EWD} & \textbf{SIR} & \textbf{KGW} & \textbf{EWD} \\
\midrule
BoolQ         & 0.91 & 0.93 & 0.70 & 0.88 & 0.93 \\
SST2          & 0.84 & 0.86 & 0.51 & 0.82 & 0.81 \\
CB            & 0.89 & 0.89 & 0.69 & 0.85 & 0.84 \\
HellaSwag     & 0.99 & 1.00 & 0.92 & 0.97 & 0.98 \\
PIQA          & 0.97 & 1.00 & 0.94 & 1.00 & 0.96 \\
SQuAD2        & 0.93 & 0.96 & 0.91 & 0.89 & 0.86 \\
DROP          & 0.86 & 0.87 & 0.69 & 0.87 & 0.85 \\
WMT14-En-Fr   & 0.93 & 0.97 & 0.85 & 0.91 & 0.93 \\
WMT20-En-De   & 0.92 & 0.95 & 0.87 & 0.91 & 0.94 \\
\midrule
Average       & 0.92 & 0.94 & 0.79 & 0.90 & 0.90 \\
\bottomrule
\end{tabular}
}
\caption{\newtext{Normalized scores for LLaMA 7B under the pareto-optimal `moderate' calibration and under $\gamma=0.5$. The effective $\gamma$ for SIR is $0.5$ by default.}}
\label{tab:why_sir_bad}
\end{table}

\newtext{Although Table~\ref{tab:why_sir_bad} shows that the normalized scores under KGW and EWD become somewhat closer to those under SIR when enforcing $\gamma=0.5$, they still remain significantly higher suggesting that the choice of $\gamma$ cannot be the only reason for SIR's underperformance.}

\newtext{\paragraph{Effect of model strength.} In Table~\ref{tab:scaling}, we show the effect of applying KGW-, EWD- and SIR-based watermarks to two stronger models in the LLaMA family: the larger LLaMA 13B and the similarly sized 7B model from the subsequent LLaMA2 generation~\citep{touvron2023llama2openfoundation}. Both stronger models show slightly larger normalized scores than LLaMA 7B, suggesting that stronger models may see smaller utility drops upon watermarking.}

\begin{table*}[ht]
\centering
\small 
\begin{tabular}{lccc||ccc|ccc}
\toprule
\textbf{Dataset} & \multicolumn{3}{c}{\textbf{LLaMA 7B}} & \multicolumn{3}{c}{\textbf{LLaMA 13B}} & \multicolumn{3}{c}{\textbf{LLaMA2 7B}} \\
\cmidrule(lr){2-4} \cmidrule(lr){5-7} \cmidrule(lr){8-10}
 & \textbf{KGW} & \textbf{EWD} & \textbf{SIR} & \textbf{KGW} & \textbf{EWD} & \textbf{SIR} & \textbf{KGW} & \textbf{EWD} & \textbf{SIR} \\
\midrule
BoolQ         & 0.91 & 0.93 & 0.70 & 0.96 & 0.97 & 0.87 & 0.96 & 0.98 & 0.90 \\
SST2          & 0.84 & 0.86 & 0.51 & 0.94 & 0.97 & 0.84 & 0.93 & 0.93 & 0.79 \\
CB            & 0.89 & 0.89 & 0.69 & 0.92 & 0.93 & 0.77 & 0.97 & 0.97 & 0.92 \\
HellaSwag     & 0.99 & 1.00 & 0.92 & 1.02 & 0.99 & 0.95 & 1.00 & 1.00 & 0.97 \\
PIQA          & 0.97 & 1.00 & 0.94 & 0.98 & 0.99 & 0.97 & 0.96 & 1.00 & 1.00 \\
SQuAD2        & 0.93 & 0.96 & 0.91 & 0.91 & 0.95 & 0.89 & 0.87 & 0.97 & 0.83 \\
DROP          & 0.86 & 0.87 & 0.69 & 0.81 & 0.72 & 0.58 & 0.79 & 0.92 & 0.69 \\
WMT14-En-Fr   & 0.93 & 0.97 & 0.85 & 0.94 & 0.97 & 0.90 & 0.95 & 0.97 & 0.92 \\
WMT20-En-De   & 0.92 & 0.95 & 0.87 & 0.95 & 0.98 & 0.93 & 0.91 & 0.99 & 0.94 \\
\midrule
Average       & 0.92 & 0.94 & 0.79 & 0.94 & 0.94 & 0.86 & 0.93 & 0.97 & 0.88 \\
\bottomrule
\end{tabular}
\caption{\newtext{Normalized scores of LLaMA 7B, LLaMA 13B and LLaMA2 7B models with KGW, EWD and SIR watermarks, along with the average scores (in the last row). Normalized scores appear slightly larger for LLaMA 13B and LLaMA2 7B (compared to LLaMA 7B), suggesting that stronger models do not see as much utility drop.}}
\label{tab:scaling}
\end{table*}

\section{Related Work}
\paragraph{Watermarking text.} Watermarking discrete-valued text data has classically been considered difficult~\citep{Petitcolas1999InformationHS, Katzenbeisser}. Early attempts involved rule-based synonym substitutions and parse-tree modifications~\citep{Chiang2003NaturalLW, topkara2006natural, Venugopal2011WatermarkingTO}. Seeing that implanting strong watermarks without severely degrading text quality was challenging for these approaches, later work utilized LSTMs~\citep{fang-etal-2017-generating} and masked language-models~\citep{Ueoka2021FrustratinglyEE} for generating watermarked text. The popularity of autoregressive LLMs has spurred fresh interest in text-watermarking techniques. \citet{pmlr-v202-kirchenbauer23a} introduced a method for implanting watermarks into LLM generations by upsampling a subset of tokens during the decoding phase. This has inspired much followup work to make LLM-watermarks robust to paraphrase attacks~\citep{kirchenbauer2023reliability, hou-etal-2024-semstamp, ren-etal-2024-robust, liu2023semanticrobust4, zhao2024provable}, encode multibit information~\citep{yoo-etal-2024-advancing, qu2024provablyrobustmultibitwatermarking, wang2024towards}, distill watermarks into standalone language models~\citep{gu2024on}, and reduce the degree of watermark-induced text degradation~\citep{wu2024dipmark, takezawa2024necessary, lu-etal-2024-entropy, chen-etal-2024-watme}. Although these works, collectively called the KGW family of watermarks are by far the most popular LLM watermarks used today, there also exist other cryptographically-inspired watermarking schemes~\citep{christ2023undetectable, aaronson2023watermarking, kuditipudi2024robust}.

\newtext{\paragraph{Downstream effects of watermarking.}
Most prior work on watermarking has evaluated their resulting models using perplexity of the generated text. \citet{pmlr-v202-kirchenbauer23a} evaluates the performance of watermarked models on a single question-answering task. Some follow-up work~\citep{fernandez2023bricks} conducts small-scale evaluations but does not attempt to uncover the causes for observed performance drops. One contemporaneous work~\citep{tu-etal-2024-waterbench} performs a similar study to ours over a broad range of tasks but in contrast to our work, chooses watermark hyperparameters relatively arbitrarily (see Section~\ref{sec:evaluation_approach}) which we believe limits the practical applicability of their findings.
To the best of our knowledge, our study is the first to conduct a principled analysis on the downstream effects of watermarking schemes over a broader spectrum of tasks, shedding light on underlying reasons for the observed trade-offs. }

\section{Conclusion}

We evaluate the extent to which watermarks from the KGW family hurt downstream performance by examining three watermark and three LLMs over a 
diverse suite of NLP tasks. We motivate a categorization of tasks into 4 buckets and analyze causes for the observed trade-offs in each category.

We find the performance trade-offs for each category vary in a manner that simple perplexity measurements cannot capture or predict (an assumption implicit in prior work). Watermarks, under realistic hyperparameters, can cause significant drops in LLMs' effective utility across all tasks. We observe drops of $10$ to $20$\% in CLS tasks in the average case, which shoot up to $100$\% in the worst case. We notice degradations of about $7$\% in MCQ tasks, $10$--$15$\% in short-form generation, and $5$--$15$\% in long-form generation tasks. \newtext{We also find some evidence that the downstream trade-offs posed by the KGW family of watermarks may diminish with increasing model strength.}

We believe that our work will (i) allow developers and practitioners to make informed choices about watermarked LLMs and their adaptations, (ii) spur research into novel watermarking strategies that present better trade-offs, and (iii) inspire techniques for maintaining model performance under existing watermarking schemes.

\section*{Limitations}

We restrict our analysis in this work to empirically evaluating the downstream performance of three representatives from the KGW family of watermarks. While we perform some analyses and give some theoretical intuitions, it may be possible to establish a concrete theoretical framework for (KGW-based) watermarked models' downstream trade-offs. We leave such analyses to future work.

Although our findings likely transfer to most KGW-based watermarks, unrelated schemes such as \citet{aaronson2023watermarking} and \citet{christ2023undetectable}, lie outside the scope of our work.

We only analyze the effect of watermarking under the typical decoding strategies used for each of the tasks. It is plausible that not all decoding strategies would be similarly affected by KGW-based watermarks. There may also exist watermark-aware decoding strategies designed to mitigate performance drops. 
This possibility presents an exciting avenue for future work.

\section*{Acknowledgments}
We are grateful to reviewers 
for their constructive feedback.
We also thank Navreet Kaur, Shashwat Singh, Saksham Rastogi, Nihar B. Shah, and Priyanka Agrawal for their insights and feedback for this work. DP acknowledges Adobe Inc., Google Research, Schmidt Sciences, National Payments Corporations of India (NPCI) and Pratiksha Trust for supporting his group’s research.

\bibliography{cite}
\bibliographystyle{acl_natbib}

\newpage
\appendix
\clearpage

\section{Watermark Hyperparameters}
We show the watermark hyperparameters we use for our main experiments (arrived at through our calibration procedure described in Section~\ref{sec:evaluation_approach}) in Table~\ref{tab:gamma_delta}.

\begin{table}[h!]
\centering
\small
\begin{tabular}{ccccccc}
\toprule
\multirow{2}{*}{} & \multicolumn{2}{c}{KGW} & \multicolumn{2}{c}{EWD} & \multicolumn{2}{c}{SIR} \\ 
Model & $\gamma$ & $\delta$ & $\gamma$ & $\delta$ & $\gamma$ & $\delta$ \\
\midrule
OPT-6.8B & 0.25 & 1.21 & 0.1 & 1.41 & 0.5 & 0.86 \\
LLaMA-7B & 0.1 & 2.13 & 0.1 & 1.72 & 0.5 & 1.31 \\
Mistral-7B & 0.1 & 1.8 & 0.1 & 1.5 & 0.5 & 1.02 \\
\bottomrule
\end{tabular}
\caption{We show the $(\gamma, \delta)$ hyperparameters obtained through the calibration procedure described in Section~\ref{sec:evaluation_approach} for each model and moderate watermark. Note that the effective $\gamma$ values for SIR are determined dynamically at each generation step, but we empirically verify that they always take values very close to $0.5$.}
\label{tab:gamma_delta}
\end{table}

\section{Additional Results}
\label{app:results}
\subsection{Results for Light, Moderate and Heavy settings}
\label{app:all_scores}
We provide evaluation results for each task, model and watermark we study in Table~\ref{tab:all_scores}.

\begin{table*}[htbp]
\centering
\tiny
\begin{tabular}{rrrccccccccc}
\toprule
\textbf{Model} & \textbf{Watermark} & \textbf{Intensity} & \textbf{BoolQ} & \textbf{SST2} & \textbf{CB} & \textbf{HellaSwag} & \textbf{PIQA} & \textbf{SQuADv2} & \textbf{DROP} & \textbf{WMT14-En-Fr} & \textbf{WMT20-En-De} \\
\midrule
OPT-6.8B & KGW & light & $0.92$ & $0.91$ & $1.11$ & $0.97$ & $0.98$ & $0.93$ & $0.98$ & $0.98$ & $0.99$ \\                                                       
 & & moderate & $0.86$ & $0.83$ & $0.81$ & $0.93$ & $0.98$ & $0.93$ & $0.97$ & $0.98$ & $0.96$ \\                                                                
 & & heavy & $0.85$ & $0.83$ & $0.80$ & $0.90$ & $0.94$ & $0.81$ & $0.96$ & $0.89$ & $0.88$ \\                                                                   
 & EWD & light & $1.15$ & $0.91$ & $0.96$ & $1.12$ & $1.06$ & $1.00$ & $0.97$ & $1.01$ & $0.97$ \\                                                               
 & & moderate & $0.92$ & $0.91$ & $0.90$ & $0.93$ & $1.00$ & $1.00$ & $0.95$ & $0.98$ & $0.96$ \\                                                                
 & & heavy & $1.04$ & $0.83$ & $0.83$ & $1.08$ & $0.94$ & $0.90$ & $0.87$ & $0.95$ & $0.91$ \\                                                                   
 & SIR & light & $0.86$ & $0.85$ & $1.06$ & $0.96$ & $1.06$ & $0.96$ & $0.98$ & $0.98$ & $0.98$ \\                                                               
 & & moderate & $0.69$ & $0.64$ & $0.70$ & $0.94$ & $1.10$ & $0.92$ & $1.01$ & $0.90$ & $0.95$ \\                                                                
 & & heavy & $0.50$ & $0.50$ & $0.56$ & $0.68$ & $0.82$ & $0.22$ & $0.65$ & $0.42$ & $0.63$ \\                                                                   
\midrule
LLaMA-7B & KGW & light & $0.93$ & $0.86$ & $0.89$ & $1.00$ & $0.97$ & $0.95$ & $0.88$ & $0.96$ & $0.93$ \\                                                       
 & & moderate & $0.91$ & $0.84$ & $0.89$ & $0.99$ & $0.97$ & $0.93$ & $0.86$ & $0.93$ & $0.92$ \\                                                                
 & & heavy & $0.86$ & $0.82$ & $0.89$ & $0.99$ & $0.97$ & $0.84$ & $0.78$ & $0.83$ & $0.76$ \\                                                                   
 & EWD & light & $0.92$ & $0.81$ & $0.90$ & $0.99$ & $1.04$ & $0.97$ & $0.94$ & $0.97$ & $0.98$ \\                                                               
 & & moderate & $0.93$ & $0.86$ & $0.89$ & $1.00$ & $1.00$ & $0.96$ & $0.87$ & $0.97$ & $0.95$ \\                                                                
 & & heavy & $0.84$ & $0.73$ & $0.89$ & $0.97$ & $0.97$ & $0.81$ & $0.78$ & $0.89$ & $0.90$ \\                                                                   
 & SIR & light & $0.91$ & $0.75$ & $0.73$ & $0.97$ & $1.06$ & $0.91$ & $0.86$ & $0.96$ & $0.99$ \\                                                               
 & & moderate & $0.70$ & $0.51$ & $0.69$ & $0.92$ & $0.94$ & $0.91$ & $0.69$ & $0.85$ & $0.87$ \\                                                                
 & & heavy & $0.50$ & $0.50$ & $0.48$ & $0.26$ & $0.42$ & $0.37$ & $0.37$ & $0.40$ & $0.46$ \\                                                                   
\midrule
Mistral-7B & KGW & light & $0.97$ & $0.98$ & $0.94$ & $1.01$ & $1.01$ & $1.01$ & $1.26$ & $0.99$ & $0.98$ \\                                                     
 & & moderate & $0.96$ & $0.97$ & $0.90$ & $0.99$ & $1.01$ & $0.97$ & $1.29$ & $0.97$ & $0.97$ \\                                                                
 & & heavy & $0.92$ & $0.95$ & $0.83$ & $0.98$ & $0.97$ & $0.96$ & $1.22$ & $0.93$ & $0.92$ \\                                                                   
 & EWD & light & $0.93$ & $0.97$ & $0.94$ & $1.06$ & $0.98$ & $0.97$ & $0.94$ & $0.99$ & $0.99$ \\                                                               
 & & moderate & $0.97$ & $0.98$ & $0.93$ & $0.99$ & $0.99$ & $0.99$ & $0.93$ & $0.98$ & $0.98$ \\                                                                
 & & heavy & $0.90$ & $0.95$ & $0.90$ & $1.03$ & $0.98$ & $0.98$ & $0.81$ & $0.96$ & $0.96$ \\                                                                   
 & SIR & light & $0.97$ & $0.98$ & $0.92$ & $1.02$ & $0.97$ & $0.96$ & $0.89$ & $0.99$ & $1.00$ \\                                                               
 & & moderate & $0.85$ & $0.90$ & $0.69$ & $0.96$ & $0.88$ & $0.90$ & $0.68$ & $0.93$ & $0.94$ \\                                                                
 & & heavy & $0.50$ & $0.50$ & $0.45$ & $0.30$ & $0.37$ & $0.44$ & $0.24$ & $0.30$ & $0.37$ \\           
\bottomrule

 \end{tabular}
 
 \caption{Normalized scores for all Model, Watermark, Intensity, Dataset combinations we evaluate.}
 \label{tab:all_scores}
 \end{table*}

\subsection{LGEN Examples}

We present 
representative   
examples for 
LGEN tasks. Examples from the WMT14-En-Fr task are tabulated in Table~\ref{tab:enfr_translation_examples} and examples from WMT20-En-De are tabulated in Table~\ref{tab:ende_translation_examples}.

\begin{table*}[htbp]
\centering
\small
\begin{tabular}{@{} p{0.15\textwidth} p{0.12\textwidth} p{0.6\textwidth} @{}}
\toprule
\textbf{Model} & & \textbf{Content} \\
\midrule
 & Input & Repainted Traffic Lights: MP Asks Brigitte Grouwels to Resign \\ 

& Ground Truth & Feux repeints: un député demande la démission de Brigitte Grouwels \textit{[Repainted lights: a deputy calls for the resignation of Brigitte Grouwels]} \\
\midrule
LLaMA 7B & Output  & Les feux de circulation ont été repeints: le député a demandé à Brigitte Grouweels de démissionner. \textit{[The traffic lights have been repainted: the MP asked Brigitte Grouweels to resign.]} \\ 
+ moderate KGW & Output  & Les feux de circulation sont rénovés, mais c'est la ministre qui se trouve sur le gril, demande à être relevée de ses fonctions. \textit{[The traffic lights are \textbf{renovated, but it is the minister who is on the grill, asking to be relieved of her duties.}]} \\

\midrule
\midrule
& Input & This really is a must for our nation. \\
& Ground Truth & Cela est vraiment indispensable pour notre nation. \textit{[This is truly essential for our nation]} \\
\midrule
Mistral 7B & Output & Cela est vraiment indispensable pour notre nation. \textit{[This is truly essential for our nation]}\\
+ moderate SIR & Output & C'est une chose qui nous manque réellement. \textit{[\textbf{This is something we really miss.}]} \\

\bottomrule
\end{tabular}
\caption{Moderate watermarks can cause increase factual errors in model outputs in the WMT14-En-Fr task. We provide the input, the gold output, the unwatermarked model's output and the output generated by a model with a moderate watermark. We also provide the English back-translation (from Google Translate) for each of these.}
\label{tab:enfr_translation_examples}
\end{table*}

\begin{table*}[htbp]
\centering
\small
\begin{tabular}{@{} p{0.15\textwidth} p{0.12\textwidth} p{0.6\textwidth} @{}}
\toprule
\textbf{Model} & & \textbf{Content} \\
\midrule

& Input & He is now facing 10 counts of murder in Canada and 16 counts of attempted murder following the attack on April 23, 2018. \\
& Ground Truth & Nach dem Attentat vom 23. April 2018 wurden in Kanada nicht weniger als 10 Anklagepunkte wegen Mordes und 16 Anklagepunkte wegen versuchten Mordes gegen ihn erhoben. \textit{[Following the assassination attempt on \textbf{April 23, 2018}, no fewer than 10 charges of murder and 16 charges of attempted murder were brought against him in Canada.]} \\
\midrule
OPT 6.8B & Output & Der Mann ist jetzt auf 10 Mordverdachtsfälle in Kanada und 16 Mordverdachtungen gegenüber der Tat am 23. April 2018. \textit{[The man is now facing 10 counts of murder in Canada and 16 counts of murder for the crime on April 23, 2018.]}\\
+ moderate EWD & Output &  \textit{[The man is now convicted of 10 counts of murder and 16 attempted murders in Canada.]} \\

\midrule \midrule

& Input & However, despite fans being eager for the upcoming release, it seems we all need to sit tight as it won't be dropping on our screens until 2021. \\
& Ground Truth & Selbst wenn seine Fans die bevorstehende Ausstrahlung der Serie kaum erwarten können, bleibt Geduld angesagt, da diese erst für 2021 angekündigt wurde. \textit{[Even if his fans can hardly wait for the upcoming broadcast of the series, patience remains as it has only been announced for 2021.]} \\
\midrule
Mistral 7B & Output & Trotzdem, obwohl Fans eifrig auf die kommende Veröffentlichung warten, scheint es so, als würden wir alle auf die Sitze sitzen müssen, bis es 2011 auf unseren Bildschirmen erscheint. \textit{[Nevertheless, although fans are eagerly awaiting the upcoming release, it seems we will all have to sit on the edge of our seats until it hits our screens in 2011.]}\\
+ moderate KGW & Output & Jedoch, obwohl Fans einschließlich mich ehrgeizig auf die nächste Veröffentlichung warten, müssen wir alle auf die nächstjährige Veröffentlichungsdatum warten. \textit{[However, although fans including myself are eagerly waiting for the next release, we all have to wait for \textbf{next year's} release date.]} \\
\bottomrule
\end{tabular}
\caption{Moderate watermarks can cause increase factual errors in model outputs in the WMT20-En-De task. We provide the input, the gold output, the unwatermarked model's output and the output generated by a model with a moderate watermark. We also provide the English back-translation (from Google Translate) for each of these.}
\label{tab:ende_translation_examples}
\end{table*}

\section{Task Evaluation Details}
\label{app:task_evaluation_details}
\subsection{Decoding}
\begin{enumerate}
    \item \textbf{CLS} tasks are tasks framed as k-class classification problems with static (and often short) labels that are common across all test examples. These are evaluated by picking the class label that the model assigns the highest probability to. Formally, the input text $\mathbf{x}$ is formatted using a suitable prompt template $T$ and the class $\mathbf{y}$ which maximizes $p(\mathbf{y} | T(\mathbf{x}))$ is chosen as the model's prediction. 
    $$\hat{y} = \argmax_{\mathbf{y} \in L} p(\mathbf{y} | T(\mathbf{x}))$$ These tasks are also typically evaluated using accuracy metrics.
    \item The \textbf{MCQ} category includes several open-book question-answering, reading-comprehension and common-sense reasoning tasks that are posed as multiple-choice question-answering tasks to language models. In these tasks, every test input $\mathbf{x}$ is associated with a set of possible answer choices $L(\mathbf{x})$. When a test input $\mathbf{x}$ is formatted using a suitable template $T$, the answer choice that the language model assigns the highest average log likelihood to,
    $$\argmax_{\mathbf{y} \in L(\mathbf{x})} \quad \text{avg-log-likelihood}(\mathbf{y} | T(\mathbf{x}))$$ is chosen as the model's prediction. These tasks are also typically evaluated using accuracy metrics.
    
    \item \textbf{SGEN} includes open-domain question-answering and reading-comprehension tasks are posed to language models as short-form conditional generation tasks and require models to output concise free-form responses. Given a test input $\mathbf{x}$ formatted using a prompt template $T$, the model produces a sequence $\mathbf{y^*}$ which maximizes the conditional likelihood $p(\mathbf{y} | T(\mathbf{x}))$.
    $$\mathbf{y}^* = \argmax_{\mathbf{y}} p(\mathbf{y} | T(\mathbf{x}))$$
    Typically, the generated sequence is bounded by a certain length or concludes when the model outputs an end-of-sequence token. The generated sequences are typically evaluated against gold sequences using F1 scores. %
    
    \item \textbf{LGEN} represents all long-form generation tasks including machine translation and summarization. For an input text $\mathbf{x}$, formatted using an appropriate prompt template $T$, the model is tasked with producing an extended sequence $\mathbf{y^*}$ that maximizes the conditional likelihood $p(\mathbf{y} | T(\mathbf{x}))$.
    $$\mathbf{y}^* = \argmax_{\mathbf{y}} p(\mathbf{y} | T(\mathbf{x}))$$
    The generated sequences are often evaluated against multiple ground truth references using metrics such as ROUGE and BLEU, allowing for some flexibility for paraphrases. %

\end{enumerate}

\end{document}